\documentclass[reprint,superscriptaddress,showpacs,amsmath,amssymb,aps,prl]{revtex4-2}

\usepackage{subfiles}
\usepackage{graphicx}
\usepackage{dcolumn}
\usepackage{bm}
\usepackage{hyperref}
\usepackage[caption=false]{subfig}
\usepackage[export]{adjustbox}
\usepackage{xcolor}

\begin{document}

\title{Spatial Analysis of Physical Reservoir Computers}

\author{Jake Love}
\affiliation{%
 Faculty of Physics, University of Duisburg-Essen, 47057 Duisburg, Germany
}%

\author{Jeroen Mulkers}
\affiliation{
DyNaMat, Department of Solid State Sciences, Ghent University, Ghent, Belgium
}

\author{Robin Msiska}
\affiliation{%
 Faculty of Physics, University of Duisburg-Essen, 47057 Duisburg, Germany
}%

\author{George Bourianoff}
\affiliation{Senior Principle Engineer, Intel Corp. (Retired)}

\author{Jonathan Leliaert}
\affiliation{
DyNaMat, Department of Solid State Sciences, Ghent University, Ghent, Belgium
}

\author{Karin Everschor-Sitte}%
\affiliation{%
Faculty of Physics and Center for Nanointegration Duisburg-Essen (CENIDE), University of Duisburg-Essen, 47057 Duisburg, Germany
}%

\date{\today}

\begin{abstract}
Physical reservoir computing is a computational framework that implements spatiotemporal information processing directly within physical systems. By exciting nonlinear dynamical systems and creating linear models from their state, we can create highly energy-efficient devices capable of solving machine learning tasks without building a modular system consisting of millions of neurons interconnected by synapses. To act as an effective reservoir, the chosen dynamical system must have two desirable properties: nonlinearity and memory. We present task agnostic spatial measures to locally measure both of these properties and exemplify them for a specific physical reservoir based upon magnetic skyrmion textures. In contrast to typical reservoir computing metrics, these metrics can be resolved spatially and in parallel from a single input signal, allowing for efficient parameter search to design efficient and high-performance reservoirs. Additionally, we show the natural trade-off between memory capacity and nonlinearity in our reservoir's behaviour, both locally and globally. Finally, by balancing the memory and nonlinearity in a reservoir, we can improve its performance for specific tasks.
\end{abstract}

\maketitle


\section{\label{sec:introduction}Introduction:\protect}
Reservoir computing (RC) is a unified supervised learning framework derived from recurrent neural networks (RNN) that utilizes dynamical systems to perform computations~\cite{lukovsevivcius2009reservoir, verstraeten2007experimental}. A reservoir computer consists of two components; a fixed nonlinear dynamical system called a reservoir and a trainable linear readout layer. Input signals are fed into the reservoir, and the system's resulting transient states are observed. The observed states, referred to as readout values, are then fed through the linear readout layer, which can be trained to solve specific problems. By separating the complex nonlinear dynamics of the reservoir from the trainable readout layer, the connecting weights can be solved inexpensively using a least-squares model, often requiring far less training data than conventional RNNs~\cite{jaeger2002tutorial}. Typically, reservoir computers are suited to pattern recognition problems dealing with temporally or spatially correlated information, such as audio or image data ~\cite{jaeger2004harnessing}.

The RC framework is not limited to mathematical models and can also be implemented with physical dynamical systems~\cite{tanaka2019recent}. In physical RC, the input signal is fed into a physical substrate via time-dependent excitations. A finite representation of the system's transient states over the excitation is then observed via experimental methods, representing the readout values. Like the non-physical case, the readout values are put through a trained linear model to perform the desired computation. By implementing reservoir computers physically, also one takes advantage of physics's inherently parallel nature and opens doors to low cost, energy-efficient computations that are not constrained by the limitations of conventional CMOS hardware,

Although one can use any physical dynamical system for reservoir computing, some are better than others. Solving complex tasks requires that the system has several fundamental properties: high dimensionality, nonlinearity, and fading memory~\cite{dambre2012information}. A diverse set of physical dynamical systems based upon optics, spintronics~\cite{nakane2019spin} and even a bucket of water~\cite{fernando2003pattern} have already been proposed to implement the reservoir computing paradigm including magnetic ~\cite{furuta2018macromagnetic} and magnetic skyrmion based systems~\cite{prychynenko2018magnetic, bourianoff2018potential, pinna2020reservoir}.

Existing literature provides global metrics to quantify the total memory capacity~\cite{jaeger2002short} and nonlinearity~\cite{dambre2012information, inubushi2017reservoir} of reservoir computers. While these global metrics are useful for the characterisation of reservoir computers, they provide limited guidance in interpreting where the required properties, memory and nonlinearity, arise within a dynamical system.

This work introduces methods that measure the memory and nonlinearity locally for physical reservoirs with spatially correlated readouts. One can use these methods to understand which physical structures lead to memory and nonlinearity in the reservoir. By creating conditions that replicate such structures, one can parameterise future reservoirs in a directed manner to optimise a balance between nonlinearity and memory as required by possible tasks.


\section{\label{sec:methods} Spatially Resolved Metrics:\protect}
We introduce two spatially resolved metrics for memory and nonlinearity. Both metrics can be calculated in parallel from the same set of input and output data and are task agnostic, hence can be applied to various input sequences. This allows one to analyse how different input sequences, such as a sequence of uniform random numbers or series generated from a Mackey-Glass equation, couple to different structures in physical reservoirs.

\subsection{Nonlinearity}

\begin{figure*}[tb]
    \subfloat[\label{fig:estimator:a}]{%
        \includegraphics[width=0.48\linewidth]{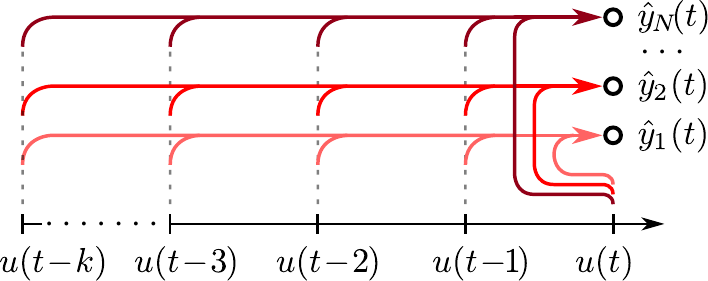}
    }\hfill
    \subfloat[\label{fig:estimator:b}]{%
        \includegraphics[width=0.48\linewidth]{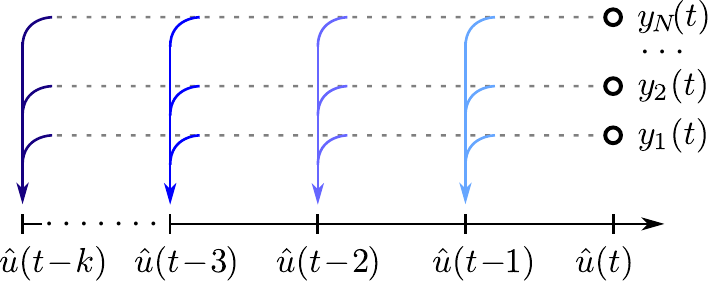}
    }\hfill
    \caption{Sketches comparing the metrics for (a) nonlinearity and (b) memory capacity. For nonlinearity, we use multiple past inputs to predict the present value of individual readout nodes. For memory capacity, we use multiple readout nodes to predict the inputs at an individual time delay in the past.}
    \label{fig:estimators}
\end{figure*}

As a reservoir is a nonlinear system with fading memory, its readout node values $\boldsymbol{y}$ can be modelled as a function of input to the system $u$ using a Volterra series~\cite{boyd1985fading} such that

\begin{equation}
    \boldsymbol{y}(t) = \boldsymbol{h}^{(0)} + \int_{\tau=0}^\infty \boldsymbol{h}^{(1)}(\tau) u(t-\tau)d\tau + \boldsymbol{y}^{\mathrm{NL}}(t).
    \label{eq:volterra}
\end{equation}
Here, $\boldsymbol{h}^{(0)}$ and $\boldsymbol{h}^{(1)}$ are zeroth and first order Volterra kernels and $\boldsymbol{y}^{\mathrm{NL}}(t)$ represents truncated nonlinear terms of the Volterra series. For a linear system, the truncated terms of Eq.~\eqref{eq:volterra} are zero and for a highly nonlinear system these truncated terms dominate. To measure the nonlinearity of each readout node in the reservoir, we quantify how important these truncated terms are when modelling the individual readout as a Volterra series.

We first model each readout node $y_n(t)$ in the reservoir as a discrete-time truncated Volterra series that can be estimated by the history of inputs $u(t)$.

\begin{equation}
    y_n^{L}(t) = h^{(0)} + \sum_{\tau=0}^\infty h^{(1)}_{\tau} u(t-\tau),
\label{eq:linear}
\end{equation}
This model is then implemented as a linear estimator of the form,

\begin{equation}
    \hat{y}_n(t) = c + \sum_{\tau=0}^k w_\tau u(t-\tau),
\label{eq:nl_estimator}
\end{equation}
where weights $w$ and constant $c$ are trained on the known input and output data, see Fig.~\ref{fig:estimator:a}. $k$ is the time delay at which we truncate the Volterra kernel, picked to be longer than the relaxation time of the reservoir. The quality of the estimator in Eq.~{\eqref{eq:nl_estimator}} is measured using the R$^2$ correlation coefficient (see Appendix~\ref{app:estimator_quality} for details) and can be considered a direct measure of how linear the response of each reservoir node is. To convert this to a measure of nonlinearity we subtract the correlation coefficient from one.

\begin{equation}
    \mathrm{NL_n} = 1 - \mathrm{R}^2[\hat{y}_n, y_n]
\end{equation}

The result is a measure of nonlinearity $\mathrm{NL_n}$ for each individual node of the reservoir bound by the interval $[0, 1]$, where zero indicates a linear relationship between reservoir input and a readout node's value, and conversely values close to one indicate a highly nonlinear relationship. The definition of nonlinearity proposed here is designed to be decoupled from the measure of memory in the system, hence it is possible to have systems with high memory and zero nonlinearity (a linear time-invariant filter) and vice versa (a high-order polynomial). This decoupling of memory and nonlinearity is in contrast to other methods such as the quadratic memory capacity task seen in Refs ~\cite{duport2012all, grigoryeva2015optimal} and the $n$-bit parity task~\cite{tsunegi2019physical}, that measure nonlinearity as a form of memory.

\subsection{Memory}
To spatially measure memory in our system, we use an adaption of the linear memory capacity metric~\cite{jaeger2002short}. In the original metric, a linear estimator is trained to recall the history of a reservoir's inputs $u$ from all the readout values of all readout nodes. We use this metric to measure memory on a per-node basis by considering the memory capacity of only the readout nodes within a threshold distance, in physical space, of the chosen node. The threshold distance and the definition of distance and the between nodes is chosen depending on the system being analysed.

For a chosen $n$th readout node we create a vector $\gamma_n$ of all other nodes of the reservoir that fall within a threshold distance, measured from $n$. We then use $\gamma_n$ to create an estimator $\hat{u}_n(t - \tau)$ that can recall the input at a time delay of $\tau$, 
\begin{equation}
    \hat{u}_n(t-\tau) = c + \sum_{i} w_i \gamma_{n,i}(t),
\label{eq:u_estimator}
\end{equation}
see Fig.~\ref{fig:estimator:b}.

The memory capacity of the nodes in $\gamma_n$ is then calculated by summing the correlation coefficient of all estimators up to a cutoff threshold, $k$, that is larger than the relaxation time of the reservoir.

\begin{equation}
    \mathrm{MC_n} = \sum_{\tau=1}^{k}\mathrm{R}^2[u(t-\tau), \hat{u_n}(t-\tau)]
\end{equation}

\subsection{Stability}
We also consider the system's stability an additional metric for physical systems as it is a prerequisite for its functional principle. Being stable is essential for physical reservoirs to ensure reproducibility and prevent the reservoir from going into a chaotic regime. For stable reservoirs, the system relaxes back to a consistent (meta)stable state in the absence of inputs.

To measure the stability, we first pass a random input random signal into the reservoir. After the input fizzles out, the reservoir is relaxed until it reaches a (meta)stable state. The difference between the system's initial and final state characterises stability, where a small (high) difference refers to a high (low) stability. The details of the difference measure can be adjusted to the specific system.

\section{\label{sec:results}Spatial analysis of Magnetic Skyrmion Reservoir:\protect}

\begin{figure*}
\begin{center}
    \includegraphics[width=.75\textwidth]{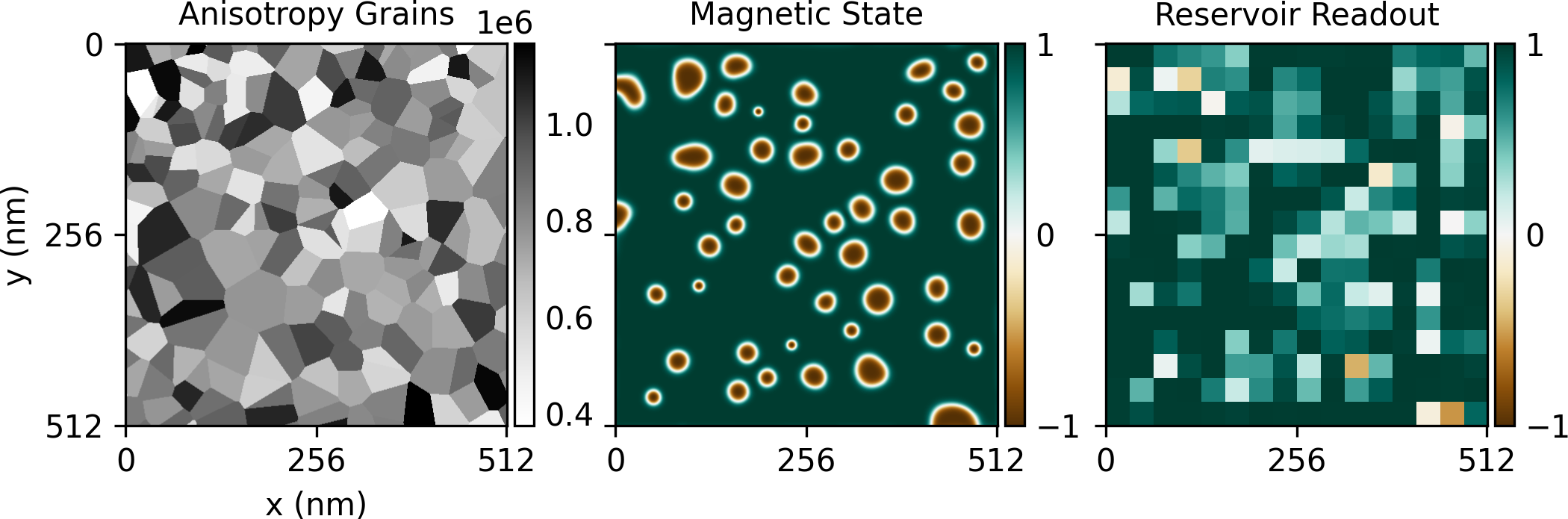}
\end{center}
\caption{Example distribution of grain inhomogeneities across a magnetic thin film. Left panel: each grain has a distinct magnetic anisotropy strength as depicted by their greyscale colour. Center panel: the non-uniform anisotropy creates pinning sites for skyrmions, resulting in their aperiodic arrangements. The color code depicts the out-of-plane component of the local normalized magnetization. Right panel: to readout values from the skyrmion reservoir, the out-of-plane projection of the magnetisation field is taken and then discretised by taking it's mean value over an area.}
\label{fig:physical_model}
\end{figure*}


To demonstrate the application of the above-introduced metrics, we apply them to a specific implementation of physical reservoir computing~\cite{pinna2020reservoir, prychynenko2018magnetic}. Magnetic skyrmions are magnetic whirls with non-trivial topology. They occur, for example, in magnetic thin films where spins favour a canted alignment, often caused by Dzyaloshinskii-Moriya interaction (DMI). Skyrmions host complex interactions with spin-polarized electric currents, defects in their host material, and also other skyrmions, resulting in deformations with highly nonlinear behaviour~\cite{bourianoff2018potential}. Skyrmions elastically deform back to their ground state upon removal of external forces; a condition which is necessary for fading memory.

In addition to the basic requirements, magnetic skyrmions are a competitive system for reservoir computing due to their nanoscale size, responsiveness to low-power excitations, and their ability to integrate with CMOS technology~\cite{finocchio2020promise} and other potential magnetic devices~\cite{torrejon2017neuromorphic, nakane2019spin, prychynenko2018magnetic, romera2018vowel}.

For our specific model, we consider an inhomogeneous magnetic thin-film with an interfacial DMI populated with magnetic skyrmions. We introduce grain inhomogeneities with varying magnetic anisotropy, resulting in a complex energy landscape. This causes specific locations on the film to be energetically favourable for skyrmions, referred to as pinning sites. As a result, the local energetic minimum state of the magnetic texture becomes an aperiodic arrangement of skyrmions, see Fig.~\ref{fig:physical_model}. 

We implement the input $u(t)$ to the reservoir as a time-varying unidirectional electric current across the film characterised by the current density $\boldsymbol{j}(t)$,

\begin{equation}
    \boldsymbol{j}(t) = j_{\mathrm{in}} u(ft) \hat{\boldsymbol{x}}.
\end{equation}
Here $\hat{\boldsymbol{x}}$ is the current direction and $j_{\mathrm{in}} u(ft)$ is the time-dependent current strength. $j_{\mathrm{in}}$ provides control over input scaling, and the factor $f$ controls time scaling. Discrete-time inputs are fed into the reservoir as a step function such that the reciprocal of $f$ is the duration of each step. 
The electrical current interacts with the magnetic skyrmions via spin-torques, resulting in a time-dependent nonlinear deformation of the magnetic texture. Provided the current is within a certain threshold, the structure will relax back to its original state when the current is removed. 

For readout nodes, we observe the changes in magnetic texture by taking a low-resolution snapshot of the film's out-of-plane magnetisation component. We choose the resolution to be of similar magnitude to the sizes of individual skyrmions as indicated in Fig.~\ref{fig:physical_model} (right panel). 

We implement the model in simulation using the MuMax3 software~\cite{vansteenkiste2014design}, with precise details on the simulation setup given in the appendix \ref{app:simulationdetails}.

\subsection{Results}

\begin{figure*}

    
    \subfloat[\label{fig:results-spatial_low}]{%
        \includegraphics[width=0.8\linewidth]{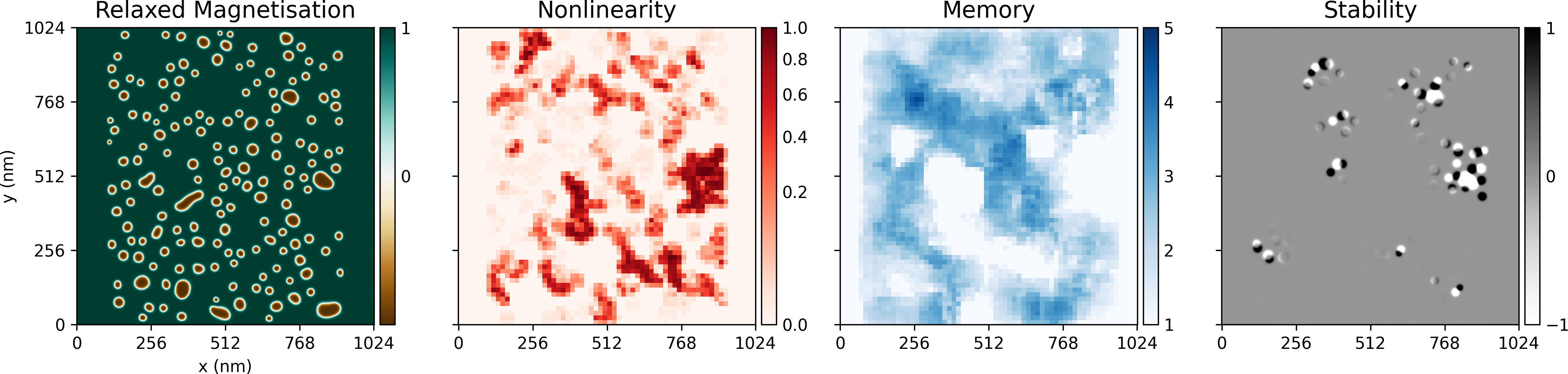}
    }\hfill

    \subfloat[\label{fig:results-spatial_high}]{%
        \includegraphics[width=0.8\linewidth]{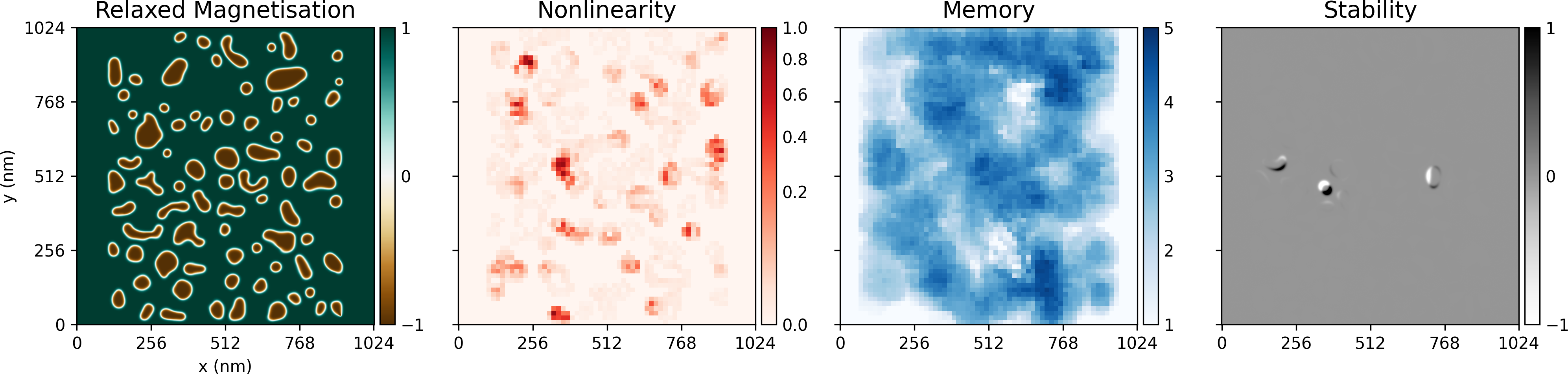}
    }\hfill
    
    \caption{Spatial analysis of two skyrmion reservoirs with (a) high DMI and (b) low DMI. The relaxed magnetisation shows the state of the reservoir when no input is applied. The nonlinearity and memory plots show the distribution said quantities over the reservoir. The stability shows the difference between the reservoir's initial and final state when measuring the nonlinearity and memory metrics.}
    \label{fig:spatial}
\end{figure*}

We first analyse two distinct skyrmion reservoirs with different DMI strengths, $3.3\times 10^{-3}$~Jm$^{-2}$ and $3.6 \times 10^{-3}$~Jm$^{-2}$, but otherwise identical dynamics. These reservoirs can be visually distinguished by the sizes of the individual skyrmion present on the thin-films as seen in the magnetisation shown in Fig~\ref{fig:spatial}. Due to the effect of DMI on the possible metastable states, the two reservoirs also have inherently different initial configurations.

Both reservoirs are tested against a discrete-time random input signal, $u(t)$. The signal has 1500 elements drawn from a random distribution with the bounds $[-1, +1]$. The input is coupled to the reservoirs with input-scaling $j_{\mathrm{in}}=300\times 10^9$~Am$^{-2}$ and time-scaling $f=4$~GHz. The memory, nonlinearity and stability are spatially analysed. The results are shown in Fig~\ref{fig:spatial}. For the case of the skyrmion reservoir, we define stability as the difference between the magnetic states before and after the random input signal has been applied.

For both reservoirs, we can observe the well-studied trade-off between memory and nonlinearity~\cite{inubushi2017reservoir, verstraeten2010memory, verstraeten2007experimental} both spatially and globally.

\begin{figure*}
    
    \includegraphics[width=\linewidth]{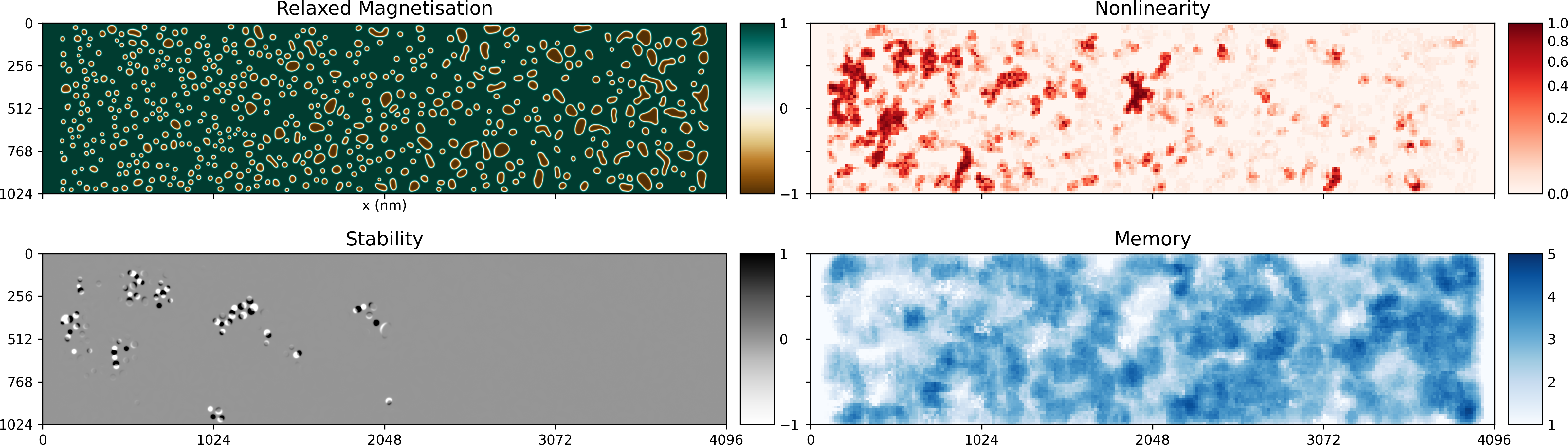}

    \caption{Resolved memory, nonlinearity and stability for system with a DMI gradient }
    
    \label{fig:spatial_grad}

\end{figure*}

\begin{figure}
    
    \includegraphics[width=\linewidth]{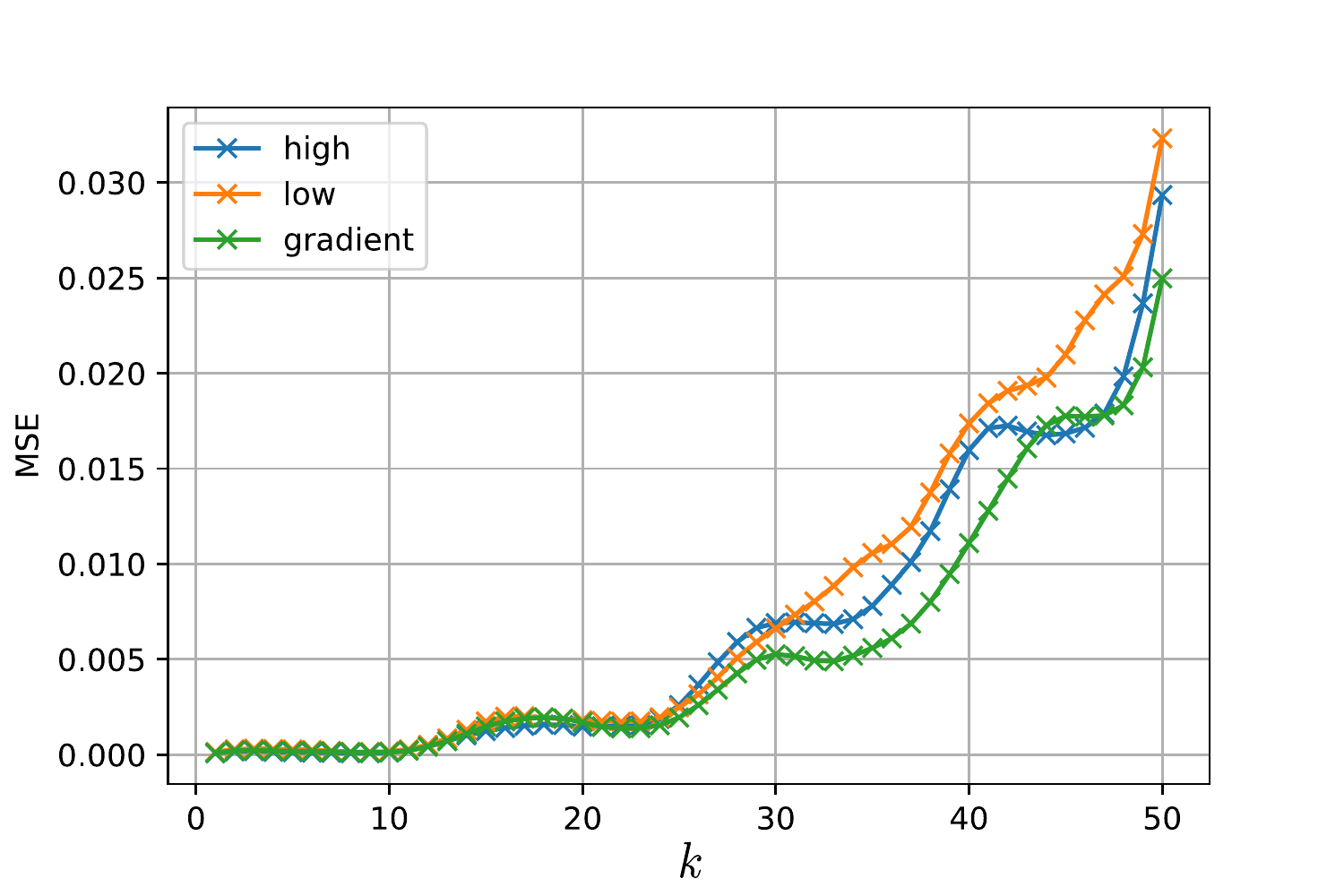}

    \caption{Mean squared error (MSE) for prediction of the Mackey-glass series for $k$ steps in the future. The results for the high, low and gradient DMI reservoirs are shown.}
    
    \label{fig:divergence}

\end{figure}

Locally, regions of high nonlinearity correspond closely to regions of low memory, as seen predominantly in Fig~\ref{fig:spatial}(a) and less so in Fig~\ref{fig:spatial}(b). However, the metrics are not entirely orthogonal as regions or with both memory and nonlinearity do coexist. Physically, these regions of high nonlinearity appear where changes to the magnetic texture are greatest. This is reflected in the stability plots where the change in initial and final states of the reservoir is observed. Although this results in the fading memory property being violated, we argue that the material can still act as a reservoir provided the global structure of the texture is maintained. In the skyrmion reservoir case, changes to local structure can be due to individual skyrmions switching between a small group of distinct pinning sites. However, this is not necessarily a problem as, once the reservoir is 'warmed up', the dynamics of the reservoir may lead to skyrmions constantly switching between pinning sites. The final pinning site upon removal of input is dependent on the local history of the input and hence, the time at which the signal is removed from the reservoir.

Globally, our analysis suggests that the high DMI reservoir has an overall higher mean memory capacity and lower mean nonlinearity for input $u(t)$ than the low DMI reservoir. As both high memory capacity and high nonlinearity are desirable for reservoir computing, we next analyse a skyrmion 'mixture reservoir' with a low to high DMI gradient. By mixing nonlinear and linear dynamics, we find that the skyrmion mixture reservoir has improved task performance over the low and high reservoirs. This complements the existing idea of an echo-state network-based 'mixture reservoir', where replacing small numbers of nonlinear nodes with linear nodes in the network improves their performance for time series forecasting tasks~\cite{inubushi2017reservoir}.

We implement the skyrmion mixture reservoir by modifying our original reservoir to possess a DMI gradient from low ($3.3\times 10^{-3}$~Jm$^{-2}$) to high ($3.6\times 10^{-3}$~Jm$^{-2}$) DMI along the axis parallel to the input current direction. We extend the reservoir geometrically along the same axis by a factor of four so that the DMI gradient is shallow. In Fig~\ref{fig:spatial_grad} we see that as a result of the DMI gradient, there is a also a gradient in the nonlinearity and memory metrics.

We next apply the reservoir to a chaotic time-series prediction task. We use a Mackey-Glass~\cite{mackey1977oscillation} system to generate a chaotic time series, see Appendix~\ref{app:mge} for details. We input this series into the reservoir and train the readout to predict the value of the series at $k$ steps in the future. The results for the prediction task are compared to the reservoirs with constant low and high DMI, extended to the same physical size as the gradient reservoir. The results of predicting up to 50 future steps are shown in Fig.~\ref{fig:divergence} in which a minor performance improvement is obtained when using the gradient DMI 'mixture reservoir'.

\section{Summary}
In this work, we have demonstrated how two independent, spatial metrics can be employed to locally quantify the properties of a reservoir computer in terms of nonlinearity and memory. Both metrics can be calculated in parallel using a small random dataset. This  allows for a rapid evaluation of reservoir models and efficient screening of reservoir parameters for 'good' reservoir behaviour and tuning reservoir performance to purpose. For the particular reservoir computing model studied, the skyrmion reservoir, we find that dynamics in which both memory and nonlinear effects are present lead to improved task performance for the Mackey-Glass series prediction task.

The metrics can also be used to better understand which environments lead to memory and nonlinearity in physical systems. By creating conditions that lead to desirable dynamics, one can parameterise reservoirs in a guided manner to optimise a balance between nonlinearity and memory. For example, in the case of the skyrmion reservoir we achieve this be introducing a gradient in the DMI parameter. The spatial metrics also provide a path to effectively choose a readout structure guided by local contributions to memory and nonlinearity. For example, having fewer effectively placed readout nodes can reduce the number of model parameters, making the engineering of such devices more feasible.

\begin{acknowledgments}
We acknowledge funding from the Emergent AI Centre funded by the Carl-Zeiss-Stiftung and the Deutsche Forschungsgemeinschaft under Project No.~320163632. J.L.~was supported by the Fonds Wetenschappelijk Onderzoek (FWO-Vlaanderen) with senior postdoctoral fellowship No12W7622N. Part of the computational resources and services used in this work were provided by the VSC (Flemish Supercomputer Center), funded by Ghent University, the Research Foundation Flanders (FWO) and the Flemish Government – department EWI. 
\end{acknowledgments}


\appendix

\section{\label{app:estimator_quality}Estimator Quality}
To measure the quality of the linear estimators for both nonlinearity, Eq.~\eqref{eq:nl_estimator} and memory Eq.~\eqref{eq:u_estimator}, we use the R$^2$ coefficient of determination. Consider the linear estimator
\begin{equation}
    \hat{y}(t) = \sum_i w_i u(t).
\end{equation}
We train the weights $w$ of the estimator using the first 75\% of our input $u(t)$ and output $y(t)$ samples, minimising the residual sum of squares error. Using the remaining 25\% of input samples, we compare the corresponding estimated values $\hat{y}(t)$ to the true value $y(t)$ with
\begin{equation}
    \mathrm{R}^2[\hat{y}, y] = \frac{\mathrm{cov}^2(\hat{y}(t), y(t))}{\sigma^2(\hat{y}(t))\sigma^2(y(t))}.
\end{equation}
Here $\sigma^2$ is the variance and 'cov' is the covariance. $\mathrm{R}^2[\hat{y}, y]$ is a measure for the estimators quality, ranging from zero (no correlation) to one (perfect prediction). In this work we use a total of 1000 data points for training and testing to measure estimator quality.

\section{\label{app:simulationdetails}Simulation Details for Skyrmion Reservoir}
The skyrmion reservoirs considered in this study were simulated using MuMax3~\cite{vansteenkiste2014design}. The homogeneous DMI reservoirs for the initial spatial analysis, were modelled with a 1024 by 1024 cell grid with a cell size of 1 nm, smaller than the exchange length of the simulated material. The gradient reservoir, along with additional homogeneous DMI reservoirs for comparing Mackey-Glass prediction performance, were modelled with a 4096 by 1024 cell grid.

We simulated the Co layer of Pt/Co/MgO~\cite{wang2018theory}, with the following material parameters: saturation magnetisation $M_{\mathrm{sat}}=580$ kAm$^{-1}$, exchange coupling constant $A_{\mathrm{ex}}=15$~pJm$^{-1}$, mean uniaxial anisotropy $K_u=0.8$~MJm$^{-3}$, temperature $T=0$~K, and we used an electrical current polarisation of 1. The DMI parameter is reservoir dependant, either $3.3\times 10^{-3}$~Jm$^{-2}$ or $3.6 \times 10^{-3}$~Jm$^{-2}$. The anisotropy grain inhomogeneities were generated using a Voronoi tessellation~\cite{Leliaert2014} and have average size of 40 nm and anisotropy variance of 20\% (see Fig.~\ref{fig:physical_model}(a) of the main text), similar to the experimentally observed grains in~\cite{bacani2019measure}. In the simulations, we do not include anisotropic magneto-resistance effects and instead assume the material resistance to be constant.

We initialise the magnetic texture by first creating a skyrmion lattice and letting it relax to a metastable state. To further stabilise the system, we excite the texture with a random input current for $200$~ns, then allow it to relax. This is to remove any low energy, initial instabilities in the reservoir and to obtain a metastable state with a higher energy barrier suitable for RC.

\section{\label{app:mge}Mackey-Glass Details}
The Mackey-Glass system is described by the equation~\cite{mackey1977oscillation}

\begin{equation}
    \frac{dx}{dt} = \frac{ax(t-\tau)}{1 + x^n(t - \tau)}-bx(t),
    \label{eq:mge}
\end{equation}
where $a=0.2$, $b=0.1$, $n=10.0$ and $\tau=23.0$. Since the dynamics of the series generated by Eq.~\ref{eq:mge} are of a different timescale to the dynamics of the reservoir, we also scale time before using it as an input. This is done such that one unit of $t$ in Eq.~\ref{eq:mge} is scaled to $1$~ns of physical input. The input is coupled to the reservoir with input-scaling $j_{\mathrm{in}}=3\times 10^{11}$ Am$^{-2}$.

\newpage
\bibliography{main.bib}

\begin{thebibliography}{27}%
\makeatletter
\providecommand \@ifxundefined [1]{%
 \@ifx{#1\undefined}
}%
\providecommand \@ifnum [1]{%
 \ifnum #1\expandafter \@firstoftwo
 \else \expandafter \@secondoftwo
 \fi
}%
\providecommand \@ifx [1]{%
 \ifx #1\expandafter \@firstoftwo
 \else \expandafter \@secondoftwo
 \fi
}%
\providecommand \natexlab [1]{#1}%
\providecommand \enquote  [1]{``#1''}%
\providecommand \bibnamefont  [1]{#1}%
\providecommand \bibfnamefont [1]{#1}%
\providecommand \citenamefont [1]{#1}%
\providecommand \href@noop [0]{\@secondoftwo}%
\providecommand \href [0]{\begingroup \@sanitize@url \@href}%
\providecommand \@href[1]{\@@startlink{#1}\@@href}%
\providecommand \@@href[1]{\endgroup#1\@@endlink}%
\providecommand \@sanitize@url [0]{\catcode `\\12\catcode `\$12\catcode
  `\&12\catcode `\#12\catcode `\^12\catcode `\_12\catcode `\%12\relax}%
\providecommand \@@startlink[1]{}%
\providecommand \@@endlink[0]{}%
\providecommand \url  [0]{\begingroup\@sanitize@url \@url }%
\providecommand \@url [1]{\endgroup\@href {#1}{\urlprefix }}%
\providecommand \urlprefix  [0]{URL }%
\providecommand \Eprint [0]{\href }%
\providecommand \doibase [0]{https://doi.org/}%
\providecommand \selectlanguage [0]{\@gobble}%
\providecommand \bibinfo  [0]{\@secondoftwo}%
\providecommand \bibfield  [0]{\@secondoftwo}%
\providecommand \translation [1]{[#1]}%
\providecommand \BibitemOpen [0]{}%
\providecommand \bibitemStop [0]{}%
\providecommand \bibitemNoStop [0]{.\EOS\space}%
\providecommand \EOS [0]{\spacefactor3000\relax}%
\providecommand \BibitemShut  [1]{\csname bibitem#1\endcsname}%
\let\auto@bib@innerbib\@empty
\bibitem [{\citenamefont {Luko{\v{s}}evi{\v{c}}ius}\ and\ \citenamefont
  {Jaeger}(2009)}]{lukovsevivcius2009reservoir}%
  \BibitemOpen
  \bibfield  {author} {\bibinfo {author} {\bibfnamefont {M.}~\bibnamefont
  {Luko{\v{s}}evi{\v{c}}ius}}\ and\ \bibinfo {author} {\bibfnamefont
  {H.}~\bibnamefont {Jaeger}},\ }\bibfield  {title} {\bibinfo {title}
  {Reservoir computing approaches to recurrent neural network training},\
  }\href@noop {} {\bibfield  {journal} {\bibinfo  {journal} {Computer Science
  Review}\ }\textbf {\bibinfo {volume} {3}},\ \bibinfo {pages} {127} (\bibinfo
  {year} {2009})}\BibitemShut {NoStop}%
\bibitem [{\citenamefont {Verstraeten}\ \emph {et~al.}(2007)\citenamefont
  {Verstraeten}, \citenamefont {Schrauwen}, \citenamefont {d’Haene},\ and\
  \citenamefont {Stroobandt}}]{verstraeten2007experimental}%
  \BibitemOpen
  \bibfield  {author} {\bibinfo {author} {\bibfnamefont {D.}~\bibnamefont
  {Verstraeten}}, \bibinfo {author} {\bibfnamefont {B.}~\bibnamefont
  {Schrauwen}}, \bibinfo {author} {\bibfnamefont {M.}~\bibnamefont
  {d’Haene}},\ and\ \bibinfo {author} {\bibfnamefont {D.}~\bibnamefont
  {Stroobandt}},\ }\bibfield  {title} {\bibinfo {title} {An experimental
  unification of reservoir computing methods},\ }\href@noop {} {\bibfield
  {journal} {\bibinfo  {journal} {Neural networks}\ }\textbf {\bibinfo {volume}
  {20}},\ \bibinfo {pages} {391} (\bibinfo {year} {2007})}\BibitemShut
  {NoStop}%
\bibitem [{\citenamefont {Jaeger}(2002{\natexlab{a}})}]{jaeger2002tutorial}%
  \BibitemOpen
  \bibfield  {author} {\bibinfo {author} {\bibfnamefont {H.}~\bibnamefont
  {Jaeger}},\ }\bibfield  {title} {\bibinfo {title} {Tutorial on training
  recurrent neural networks, covering {BPPT, RTRL, EKF} and the "echo state
  network" approach}\ }(\bibinfo  {publisher} {GMD - German National Research
  Institute for Computer Science},\ \bibinfo {year} {2002})\BibitemShut
  {NoStop}%
\bibitem [{\citenamefont {Jaeger}\ and\ \citenamefont
  {Haas}(2004)}]{jaeger2004harnessing}%
  \BibitemOpen
  \bibfield  {author} {\bibinfo {author} {\bibfnamefont {H.}~\bibnamefont
  {Jaeger}}\ and\ \bibinfo {author} {\bibfnamefont {H.}~\bibnamefont {Haas}},\
  }\bibfield  {title} {\bibinfo {title} {Harnessing nonlinearity: Predicting
  chaotic systems and saving energy in wireless communication},\ }\href@noop {}
  {\bibfield  {journal} {\bibinfo  {journal} {Science}\ } (\bibinfo {year}
  {2004})}\BibitemShut {NoStop}%
\bibitem [{\citenamefont {Tanaka}\ \emph {et~al.}(2019)\citenamefont {Tanaka},
  \citenamefont {Yamane}, \citenamefont {H{\'e}roux}, \citenamefont {Nakane},
  \citenamefont {Kanazawa}, \citenamefont {Takeda}, \citenamefont {Numata},
  \citenamefont {Nakano},\ and\ \citenamefont {Hirose}}]{tanaka2019recent}%
  \BibitemOpen
  \bibfield  {author} {\bibinfo {author} {\bibfnamefont {G.}~\bibnamefont
  {Tanaka}}, \bibinfo {author} {\bibfnamefont {T.}~\bibnamefont {Yamane}},
  \bibinfo {author} {\bibfnamefont {J.~B.}\ \bibnamefont {H{\'e}roux}},
  \bibinfo {author} {\bibfnamefont {R.}~\bibnamefont {Nakane}}, \bibinfo
  {author} {\bibfnamefont {N.}~\bibnamefont {Kanazawa}}, \bibinfo {author}
  {\bibfnamefont {S.}~\bibnamefont {Takeda}}, \bibinfo {author} {\bibfnamefont
  {H.}~\bibnamefont {Numata}}, \bibinfo {author} {\bibfnamefont
  {D.}~\bibnamefont {Nakano}},\ and\ \bibinfo {author} {\bibfnamefont
  {A.}~\bibnamefont {Hirose}},\ }\bibfield  {title} {\bibinfo {title} {Recent
  advances in physical reservoir computing: A review},\ }\href@noop {}
  {\bibfield  {journal} {\bibinfo  {journal} {Neural Networks}\ }\textbf
  {\bibinfo {volume} {115}},\ \bibinfo {pages} {100} (\bibinfo {year}
  {2019})}\BibitemShut {NoStop}%
\bibitem [{\citenamefont {Dambre}\ \emph {et~al.}(2012)\citenamefont {Dambre},
  \citenamefont {Verstraeten}, \citenamefont {Schrauwen},\ and\ \citenamefont
  {Massar}}]{dambre2012information}%
  \BibitemOpen
  \bibfield  {author} {\bibinfo {author} {\bibfnamefont {J.}~\bibnamefont
  {Dambre}}, \bibinfo {author} {\bibfnamefont {D.}~\bibnamefont {Verstraeten}},
  \bibinfo {author} {\bibfnamefont {B.}~\bibnamefont {Schrauwen}},\ and\
  \bibinfo {author} {\bibfnamefont {S.}~\bibnamefont {Massar}},\ }\bibfield
  {title} {\bibinfo {title} {Information processing capacity of dynamical
  systems},\ }\href@noop {} {\bibfield  {journal} {\bibinfo  {journal}
  {Scientific Reports}\ }\textbf {\bibinfo {volume} {2}},\ \bibinfo {pages} {1}
  (\bibinfo {year} {2012})}\BibitemShut {NoStop}%
\bibitem [{\citenamefont {Nakane}\ \emph {et~al.}(2019)\citenamefont {Nakane},
  \citenamefont {Tanaka},\ and\ \citenamefont {Hirose}}]{nakane2019spin}%
  \BibitemOpen
  \bibfield  {author} {\bibinfo {author} {\bibfnamefont {R.}~\bibnamefont
  {Nakane}}, \bibinfo {author} {\bibfnamefont {G.}~\bibnamefont {Tanaka}},\
  and\ \bibinfo {author} {\bibfnamefont {A.}~\bibnamefont {Hirose}},\
  }\bibfield  {title} {\bibinfo {title} {In a spin-wave reservoir for machine
  learning},\ }in\ \href@noop {} {\emph {\bibinfo {booktitle} {2019
  International Joint Conference on Neural Networks {IJCNN}}}}\ (\bibinfo
  {organization} {IEEE},\ \bibinfo {year} {2019})\ pp.\ \bibinfo {pages}
  {1--9}\BibitemShut {NoStop}%
\bibitem [{\citenamefont {Fernando}\ and\ \citenamefont
  {Sojakka}(2003)}]{fernando2003pattern}%
  \BibitemOpen
  \bibfield  {author} {\bibinfo {author} {\bibfnamefont {C.}~\bibnamefont
  {Fernando}}\ and\ \bibinfo {author} {\bibfnamefont {S.}~\bibnamefont
  {Sojakka}},\ }\bibfield  {title} {\bibinfo {title} {Pattern recognition in a
  bucket},\ }in\ \href@noop {} {\emph {\bibinfo {booktitle} {European
  conference on artificial life}}}\ (\bibinfo {organization} {Springer},\
  \bibinfo {year} {2003})\ pp.\ \bibinfo {pages} {588--597}\BibitemShut
  {NoStop}%
\bibitem [{\citenamefont {Furuta}\ \emph {et~al.}(2018)\citenamefont {Furuta},
  \citenamefont {Fujii}, \citenamefont {Nakajima}, \citenamefont {Tsunegi},
  \citenamefont {Kubota}, \citenamefont {Suzuki},\ and\ \citenamefont
  {Miwa}}]{furuta2018macromagnetic}%
  \BibitemOpen
  \bibfield  {author} {\bibinfo {author} {\bibfnamefont {T.}~\bibnamefont
  {Furuta}}, \bibinfo {author} {\bibfnamefont {K.}~\bibnamefont {Fujii}},
  \bibinfo {author} {\bibfnamefont {K.}~\bibnamefont {Nakajima}}, \bibinfo
  {author} {\bibfnamefont {S.}~\bibnamefont {Tsunegi}}, \bibinfo {author}
  {\bibfnamefont {H.}~\bibnamefont {Kubota}}, \bibinfo {author} {\bibfnamefont
  {Y.}~\bibnamefont {Suzuki}},\ and\ \bibinfo {author} {\bibfnamefont
  {S.}~\bibnamefont {Miwa}},\ }\bibfield  {title} {\bibinfo {title}
  {Macromagnetic simulation for reservoir computing utilizing spin dynamics in
  magnetic tunnel junctions},\ }\href@noop {} {\bibfield  {journal} {\bibinfo
  {journal} {Physical Review Applied}\ }\textbf {\bibinfo {volume} {10}},\
  \bibinfo {pages} {034063} (\bibinfo {year} {2018})}\BibitemShut {NoStop}%
\bibitem [{\citenamefont {Prychynenko}\ \emph {et~al.}(2018)\citenamefont
  {Prychynenko}, \citenamefont {Sitte}, \citenamefont {Litzius}, \citenamefont
  {Kr{\"u}ger}, \citenamefont {Bourianoff}, \citenamefont {Kl{\"a}ui},
  \citenamefont {Sinova},\ and\ \citenamefont
  {Everschor-Sitte}}]{prychynenko2018magnetic}%
  \BibitemOpen
  \bibfield  {author} {\bibinfo {author} {\bibfnamefont {D.}~\bibnamefont
  {Prychynenko}}, \bibinfo {author} {\bibfnamefont {M.}~\bibnamefont {Sitte}},
  \bibinfo {author} {\bibfnamefont {K.}~\bibnamefont {Litzius}}, \bibinfo
  {author} {\bibfnamefont {B.}~\bibnamefont {Kr{\"u}ger}}, \bibinfo {author}
  {\bibfnamefont {G.}~\bibnamefont {Bourianoff}}, \bibinfo {author}
  {\bibfnamefont {M.}~\bibnamefont {Kl{\"a}ui}}, \bibinfo {author}
  {\bibfnamefont {J.}~\bibnamefont {Sinova}},\ and\ \bibinfo {author}
  {\bibfnamefont {K.}~\bibnamefont {Everschor-Sitte}},\ }\bibfield  {title}
  {\bibinfo {title} {Magnetic skyrmion as a nonlinear resistive element: a
  potential building block for reservoir computing},\ }\href@noop {} {\bibfield
   {journal} {\bibinfo  {journal} {Physical Review Applied}\ }\textbf {\bibinfo
  {volume} {9}},\ \bibinfo {pages} {014034} (\bibinfo {year}
  {2018})}\BibitemShut {NoStop}%
\bibitem [{\citenamefont {Bourianoff}\ \emph {et~al.}(2018)\citenamefont
  {Bourianoff}, \citenamefont {Pinna}, \citenamefont {Sitte},\ and\
  \citenamefont {Everschor-Sitte}}]{bourianoff2018potential}%
  \BibitemOpen
  \bibfield  {author} {\bibinfo {author} {\bibfnamefont {G.}~\bibnamefont
  {Bourianoff}}, \bibinfo {author} {\bibfnamefont {D.}~\bibnamefont {Pinna}},
  \bibinfo {author} {\bibfnamefont {M.}~\bibnamefont {Sitte}},\ and\ \bibinfo
  {author} {\bibfnamefont {K.}~\bibnamefont {Everschor-Sitte}},\ }\bibfield
  {title} {\bibinfo {title} {Potential implementation of reservoir computing
  models based on magnetic skyrmions},\ }\href@noop {} {\bibfield  {journal}
  {\bibinfo  {journal} {{AIP} Advances}\ }\textbf {\bibinfo {volume} {8}},\
  \bibinfo {pages} {055602} (\bibinfo {year} {2018})}\BibitemShut {NoStop}%
\bibitem [{\citenamefont {Pinna}\ \emph {et~al.}(2020)\citenamefont {Pinna},
  \citenamefont {Bourianoff},\ and\ \citenamefont
  {Everschor-Sitte}}]{pinna2020reservoir}%
  \BibitemOpen
  \bibfield  {author} {\bibinfo {author} {\bibfnamefont {D.}~\bibnamefont
  {Pinna}}, \bibinfo {author} {\bibfnamefont {G.}~\bibnamefont {Bourianoff}},\
  and\ \bibinfo {author} {\bibfnamefont {K.}~\bibnamefont {Everschor-Sitte}},\
  }\bibfield  {title} {\bibinfo {title} {Reservoir computing with random
  skyrmion textures},\ }\href@noop {} {\bibfield  {journal} {\bibinfo
  {journal} {Physical Review Applied}\ }\textbf {\bibinfo {volume} {14}},\
  \bibinfo {pages} {054020} (\bibinfo {year} {2020})}\BibitemShut {NoStop}%
\bibitem [{\citenamefont {Jaeger}(2002{\natexlab{b}})}]{jaeger2002short}%
  \BibitemOpen
  \bibfield  {author} {\bibinfo {author} {\bibfnamefont {H.}~\bibnamefont
  {Jaeger}},\ }\bibfield  {title} {\bibinfo {title} {Short term memory in echo
  state networks},\ }in\ \href@noop {} {\emph {\bibinfo {booktitle} {{GMD}
  Technical Report 152}}}\ (\bibinfo {organization} {GMD - German National
  Research Institute for Computer Science},\ \bibinfo {year}
  {2002})\BibitemShut {NoStop}%
\bibitem [{\citenamefont {Inubushi}\ and\ \citenamefont
  {Yoshimura}(2017)}]{inubushi2017reservoir}%
  \BibitemOpen
  \bibfield  {author} {\bibinfo {author} {\bibfnamefont {M.}~\bibnamefont
  {Inubushi}}\ and\ \bibinfo {author} {\bibfnamefont {K.}~\bibnamefont
  {Yoshimura}},\ }\bibfield  {title} {\bibinfo {title} {Reservoir computing
  beyond memory-nonlinearity trade-off},\ }\href@noop {} {\bibfield  {journal}
  {\bibinfo  {journal} {Scientific Reports}\ }\textbf {\bibinfo {volume} {7}},\
  \bibinfo {pages} {1} (\bibinfo {year} {2017})}\BibitemShut {NoStop}%
\bibitem [{\citenamefont {Boyd}\ and\ \citenamefont
  {Chua}(1985)}]{boyd1985fading}%
  \BibitemOpen
  \bibfield  {author} {\bibinfo {author} {\bibfnamefont {S.}~\bibnamefont
  {Boyd}}\ and\ \bibinfo {author} {\bibfnamefont {L.}~\bibnamefont {Chua}},\
  }\bibfield  {title} {\bibinfo {title} {Fading memory and the problem of
  approximating nonlinear operators with volterra series},\ }\href@noop {}
  {\bibfield  {journal} {\bibinfo  {journal} {IEEE Transactions on Circuits and
  Systems}\ }\textbf {\bibinfo {volume} {32}},\ \bibinfo {pages} {1150}
  (\bibinfo {year} {1985})}\BibitemShut {NoStop}%
\bibitem [{\citenamefont {Duport}\ \emph {et~al.}(2012)\citenamefont {Duport},
  \citenamefont {Schneider}, \citenamefont {Smerieri}, \citenamefont
  {Haelterman},\ and\ \citenamefont {Massar}}]{duport2012all}%
  \BibitemOpen
  \bibfield  {author} {\bibinfo {author} {\bibfnamefont {F.}~\bibnamefont
  {Duport}}, \bibinfo {author} {\bibfnamefont {B.}~\bibnamefont {Schneider}},
  \bibinfo {author} {\bibfnamefont {A.}~\bibnamefont {Smerieri}}, \bibinfo
  {author} {\bibfnamefont {M.}~\bibnamefont {Haelterman}},\ and\ \bibinfo
  {author} {\bibfnamefont {S.}~\bibnamefont {Massar}},\ }\bibfield  {title}
  {\bibinfo {title} {All-optical reservoir computing},\ }\href@noop {}
  {\bibfield  {journal} {\bibinfo  {journal} {Optics Express}\ }\textbf
  {\bibinfo {volume} {20}},\ \bibinfo {pages} {22783} (\bibinfo {year}
  {2012})}\BibitemShut {NoStop}%
\bibitem [{\citenamefont {Grigoryeva}\ \emph {et~al.}(2015)\citenamefont
  {Grigoryeva}, \citenamefont {Henriques}, \citenamefont {Larger},\ and\
  \citenamefont {Ortega}}]{grigoryeva2015optimal}%
  \BibitemOpen
  \bibfield  {author} {\bibinfo {author} {\bibfnamefont {L.}~\bibnamefont
  {Grigoryeva}}, \bibinfo {author} {\bibfnamefont {J.}~\bibnamefont
  {Henriques}}, \bibinfo {author} {\bibfnamefont {L.}~\bibnamefont {Larger}},\
  and\ \bibinfo {author} {\bibfnamefont {J.-P.}\ \bibnamefont {Ortega}},\
  }\bibfield  {title} {\bibinfo {title} {Optimal nonlinear information
  processing capacity in delay-based reservoir computers},\ }\href@noop {}
  {\bibfield  {journal} {\bibinfo  {journal} {Scientific Reports}\ }\textbf
  {\bibinfo {volume} {5}},\ \bibinfo {pages} {1} (\bibinfo {year}
  {2015})}\BibitemShut {NoStop}%
\bibitem [{\citenamefont {Tsunegi}\ \emph {et~al.}(2019)\citenamefont
  {Tsunegi}, \citenamefont {Taniguchi}, \citenamefont {Nakajima}, \citenamefont
  {Miwa}, \citenamefont {Yakushiji}, \citenamefont {Fukushima}, \citenamefont
  {Yuasa},\ and\ \citenamefont {Kubota}}]{tsunegi2019physical}%
  \BibitemOpen
  \bibfield  {author} {\bibinfo {author} {\bibfnamefont {S.}~\bibnamefont
  {Tsunegi}}, \bibinfo {author} {\bibfnamefont {T.}~\bibnamefont {Taniguchi}},
  \bibinfo {author} {\bibfnamefont {K.}~\bibnamefont {Nakajima}}, \bibinfo
  {author} {\bibfnamefont {S.}~\bibnamefont {Miwa}}, \bibinfo {author}
  {\bibfnamefont {K.}~\bibnamefont {Yakushiji}}, \bibinfo {author}
  {\bibfnamefont {A.}~\bibnamefont {Fukushima}}, \bibinfo {author}
  {\bibfnamefont {S.}~\bibnamefont {Yuasa}},\ and\ \bibinfo {author}
  {\bibfnamefont {H.}~\bibnamefont {Kubota}},\ }\bibfield  {title} {\bibinfo
  {title} {Physical reservoir computing based on spin torque oscillator with
  forced synchronization},\ }\href@noop {} {\bibfield  {journal} {\bibinfo
  {journal} {Applied Physics Letters}\ }\textbf {\bibinfo {volume} {114}},\
  \bibinfo {pages} {164101} (\bibinfo {year} {2019})}\BibitemShut {NoStop}%
\bibitem [{\citenamefont {Finocchio}\ \emph {et~al.}(2020)\citenamefont
  {Finocchio}, \citenamefont {Di~Ventra}, \citenamefont {Camsari},
  \citenamefont {Everschor-Sitte}, \citenamefont {Amiri},\ and\ \citenamefont
  {Zeng}}]{finocchio2020promise}%
  \BibitemOpen
  \bibfield  {author} {\bibinfo {author} {\bibfnamefont {G.}~\bibnamefont
  {Finocchio}}, \bibinfo {author} {\bibfnamefont {M.}~\bibnamefont
  {Di~Ventra}}, \bibinfo {author} {\bibfnamefont {K.~Y.}\ \bibnamefont
  {Camsari}}, \bibinfo {author} {\bibfnamefont {K.}~\bibnamefont
  {Everschor-Sitte}}, \bibinfo {author} {\bibfnamefont {P.~K.}\ \bibnamefont
  {Amiri}},\ and\ \bibinfo {author} {\bibfnamefont {Z.}~\bibnamefont {Zeng}},\
  }\bibfield  {title} {\bibinfo {title} {The promise of spintronics for
  unconventional computing},\ }\href@noop {} {\bibfield  {journal} {\bibinfo
  {journal} {Journal of Magnetism and Magnetic Materials}\ ,\ \bibinfo {pages}
  {167506}} (\bibinfo {year} {2020})}\BibitemShut {NoStop}%
\bibitem [{\citenamefont {Torrejon}\ \emph {et~al.}(2017)\citenamefont
  {Torrejon}, \citenamefont {Riou}, \citenamefont {Araujo}, \citenamefont
  {Tsunegi}, \citenamefont {Khalsa}, \citenamefont {Querlioz}, \citenamefont
  {Bortolotti}, \citenamefont {Cros}, \citenamefont {Yakushiji}, \citenamefont
  {Fukushima} \emph {et~al.}}]{torrejon2017neuromorphic}%
  \BibitemOpen
  \bibfield  {author} {\bibinfo {author} {\bibfnamefont {J.}~\bibnamefont
  {Torrejon}}, \bibinfo {author} {\bibfnamefont {M.}~\bibnamefont {Riou}},
  \bibinfo {author} {\bibfnamefont {F.~A.}\ \bibnamefont {Araujo}}, \bibinfo
  {author} {\bibfnamefont {S.}~\bibnamefont {Tsunegi}}, \bibinfo {author}
  {\bibfnamefont {G.}~\bibnamefont {Khalsa}}, \bibinfo {author} {\bibfnamefont
  {D.}~\bibnamefont {Querlioz}}, \bibinfo {author} {\bibfnamefont
  {P.}~\bibnamefont {Bortolotti}}, \bibinfo {author} {\bibfnamefont
  {V.}~\bibnamefont {Cros}}, \bibinfo {author} {\bibfnamefont {K.}~\bibnamefont
  {Yakushiji}}, \bibinfo {author} {\bibfnamefont {A.}~\bibnamefont
  {Fukushima}}, \emph {et~al.},\ }\bibfield  {title} {\bibinfo {title}
  {Neuromorphic computing with nanoscale spintronic oscillators},\ }\href@noop
  {} {\bibfield  {journal} {\bibinfo  {journal} {Nature}\ }\textbf {\bibinfo
  {volume} {547}},\ \bibinfo {pages} {428} (\bibinfo {year}
  {2017})}\BibitemShut {NoStop}%
\bibitem [{\citenamefont {Romera}\ \emph {et~al.}(2018)\citenamefont {Romera},
  \citenamefont {Talatchian}, \citenamefont {Tsunegi}, \citenamefont {Araujo},
  \citenamefont {Cros}, \citenamefont {Bortolotti}, \citenamefont {Trastoy},
  \citenamefont {Yakushiji}, \citenamefont {Fukushima}, \citenamefont {Kubota}
  \emph {et~al.}}]{romera2018vowel}%
  \BibitemOpen
  \bibfield  {author} {\bibinfo {author} {\bibfnamefont {M.}~\bibnamefont
  {Romera}}, \bibinfo {author} {\bibfnamefont {P.}~\bibnamefont {Talatchian}},
  \bibinfo {author} {\bibfnamefont {S.}~\bibnamefont {Tsunegi}}, \bibinfo
  {author} {\bibfnamefont {F.~A.}\ \bibnamefont {Araujo}}, \bibinfo {author}
  {\bibfnamefont {V.}~\bibnamefont {Cros}}, \bibinfo {author} {\bibfnamefont
  {P.}~\bibnamefont {Bortolotti}}, \bibinfo {author} {\bibfnamefont
  {J.}~\bibnamefont {Trastoy}}, \bibinfo {author} {\bibfnamefont
  {K.}~\bibnamefont {Yakushiji}}, \bibinfo {author} {\bibfnamefont
  {A.}~\bibnamefont {Fukushima}}, \bibinfo {author} {\bibfnamefont
  {H.}~\bibnamefont {Kubota}}, \emph {et~al.},\ }\bibfield  {title} {\bibinfo
  {title} {Vowel recognition with four coupled spin-torque nano-oscillators},\
  }\href@noop {} {\bibfield  {journal} {\bibinfo  {journal} {Nature}\ }\textbf
  {\bibinfo {volume} {563}},\ \bibinfo {pages} {230} (\bibinfo {year}
  {2018})}\BibitemShut {NoStop}%
\bibitem [{\citenamefont {Vansteenkiste}\ \emph {et~al.}(2014)\citenamefont
  {Vansteenkiste}, \citenamefont {Leliaert}, \citenamefont {Dvornik},
  \citenamefont {Helsen}, \citenamefont {Garcia-Sanchez},\ and\ \citenamefont
  {Van~Waeyenberge}}]{vansteenkiste2014design}%
  \BibitemOpen
  \bibfield  {author} {\bibinfo {author} {\bibfnamefont {A.}~\bibnamefont
  {Vansteenkiste}}, \bibinfo {author} {\bibfnamefont {J.}~\bibnamefont
  {Leliaert}}, \bibinfo {author} {\bibfnamefont {M.}~\bibnamefont {Dvornik}},
  \bibinfo {author} {\bibfnamefont {M.}~\bibnamefont {Helsen}}, \bibinfo
  {author} {\bibfnamefont {F.}~\bibnamefont {Garcia-Sanchez}},\ and\ \bibinfo
  {author} {\bibfnamefont {B.}~\bibnamefont {Van~Waeyenberge}},\ }\bibfield
  {title} {\bibinfo {title} {The design and verification of mumax3},\
  }\href@noop {} {\bibfield  {journal} {\bibinfo  {journal} {{AIP} advances}\
  }\textbf {\bibinfo {volume} {4}},\ \bibinfo {pages} {107133} (\bibinfo {year}
  {2014})}\BibitemShut {NoStop}%
\bibitem [{\citenamefont {Verstraeten}\ \emph {et~al.}(2010)\citenamefont
  {Verstraeten}, \citenamefont {Dambre}, \citenamefont {Dutoit},\ and\
  \citenamefont {Schrauwen}}]{verstraeten2010memory}%
  \BibitemOpen
  \bibfield  {author} {\bibinfo {author} {\bibfnamefont {D.}~\bibnamefont
  {Verstraeten}}, \bibinfo {author} {\bibfnamefont {J.}~\bibnamefont {Dambre}},
  \bibinfo {author} {\bibfnamefont {X.}~\bibnamefont {Dutoit}},\ and\ \bibinfo
  {author} {\bibfnamefont {B.}~\bibnamefont {Schrauwen}},\ }\bibfield  {title}
  {\bibinfo {title} {Memory versus non-linearity in reservoirs},\ }in\
  \href@noop {} {\emph {\bibinfo {booktitle} {The 2010 international joint
  conference on neural networks {IJCNN}}}}\ (\bibinfo {organization} {IEEE},\
  \bibinfo {year} {2010})\ pp.\ \bibinfo {pages} {1--8}\BibitemShut {NoStop}%
\bibitem [{\citenamefont {Mackey}\ and\ \citenamefont
  {Glass}(1977)}]{mackey1977oscillation}%
  \BibitemOpen
  \bibfield  {author} {\bibinfo {author} {\bibfnamefont {M.~C.}\ \bibnamefont
  {Mackey}}\ and\ \bibinfo {author} {\bibfnamefont {L.}~\bibnamefont {Glass}},\
  }\bibfield  {title} {\bibinfo {title} {Oscillation and chaos in physiological
  control systems},\ }\href@noop {} {\bibfield  {journal} {\bibinfo  {journal}
  {Science}\ }\textbf {\bibinfo {volume} {197}},\ \bibinfo {pages} {287}
  (\bibinfo {year} {1977})}\BibitemShut {NoStop}%
\bibitem [{\citenamefont {Wang}\ \emph {et~al.}(2018)\citenamefont {Wang},
  \citenamefont {Yuan},\ and\ \citenamefont {Wang}}]{wang2018theory}%
  \BibitemOpen
  \bibfield  {author} {\bibinfo {author} {\bibfnamefont {X.}~\bibnamefont
  {Wang}}, \bibinfo {author} {\bibfnamefont {H.}~\bibnamefont {Yuan}},\ and\
  \bibinfo {author} {\bibfnamefont {X.}~\bibnamefont {Wang}},\ }\bibfield
  {title} {\bibinfo {title} {A theory on skyrmion size},\ }\href@noop {}
  {\bibfield  {journal} {\bibinfo  {journal} {Communications Physics}\ }\textbf
  {\bibinfo {volume} {1}},\ \bibinfo {pages} {1} (\bibinfo {year}
  {2018})}\BibitemShut {NoStop}%
\bibitem [{\citenamefont {Leliaert}\ \emph {et~al.}(2014)\citenamefont
  {Leliaert}, \citenamefont {Van~de Wiele}, \citenamefont {Vansteenkiste},
  \citenamefont {Laurson}, \citenamefont {Durin}, \citenamefont {Dupré},\ and\
  \citenamefont {Van~Waeyenberge}}]{Leliaert2014}%
  \BibitemOpen
  \bibfield  {author} {\bibinfo {author} {\bibfnamefont {J.}~\bibnamefont
  {Leliaert}}, \bibinfo {author} {\bibfnamefont {B.}~\bibnamefont {Van~de
  Wiele}}, \bibinfo {author} {\bibfnamefont {A.}~\bibnamefont {Vansteenkiste}},
  \bibinfo {author} {\bibfnamefont {L.}~\bibnamefont {Laurson}}, \bibinfo
  {author} {\bibfnamefont {G.}~\bibnamefont {Durin}}, \bibinfo {author}
  {\bibfnamefont {L.}~\bibnamefont {Dupré}},\ and\ \bibinfo {author}
  {\bibfnamefont {B.}~\bibnamefont {Van~Waeyenberge}},\ }\bibfield  {title}
  {\bibinfo {title} {Current-driven domain wall mobility in polycrystalline
  permalloy nanowires: A numerical study},\ }\href@noop {} {\bibfield
  {journal} {\bibinfo  {journal} {Journal of Applied Physics}\ }\textbf
  {\bibinfo {volume} {115}},\ \bibinfo {pages} {233903} (\bibinfo {year}
  {2014})}\BibitemShut {NoStop}%
\bibitem [{\citenamefont {Ba{\'c}ani}\ \emph {et~al.}(2019)\citenamefont
  {Ba{\'c}ani}, \citenamefont {Marioni}, \citenamefont {Schwenk},\ and\
  \citenamefont {Hug}}]{bacani2019measure}%
  \BibitemOpen
  \bibfield  {author} {\bibinfo {author} {\bibfnamefont {M.}~\bibnamefont
  {Ba{\'c}ani}}, \bibinfo {author} {\bibfnamefont {M.~A.}\ \bibnamefont
  {Marioni}}, \bibinfo {author} {\bibfnamefont {J.}~\bibnamefont {Schwenk}},\
  and\ \bibinfo {author} {\bibfnamefont {H.~J.}\ \bibnamefont {Hug}},\
  }\bibfield  {title} {\bibinfo {title} {How to measure the local
  dzyaloshinskii-moriya interaction in skyrmion thin-film multilayers},\
  }\href@noop {} {\bibfield  {journal} {\bibinfo  {journal} {Scientific
  Reports}\ }\textbf {\bibinfo {volume} {9}},\ \bibinfo {pages} {1} (\bibinfo
  {year} {2019})}\BibitemShut {NoStop}%
\end{thebibliography}%

\end{document}